\documentclass[twoside]{article}

\usepackage[accepted]{aistats2022}
\usepackage{graphicx}
\usepackage{algorithm}
\usepackage{algorithmicx}
\usepackage[noend]{algpseudocode}
\usepackage{subcaption}
\usepackage{enumitem}
\usepackage{mathtools}
\usepackage[round]{natbib}
\DeclareMathOperator*{\argmax}{argmax}

\newcommand\blfootnote[1]{%
	\begingroup
	\renewcommand\thefootnote{}\footnote{#1}%
	\addtocounter{footnote}{-1}%
	\endgroup
}

\begin{document}

%

%

\twocolumn[

\aistatstitle{Decentralized EM to Learn Gaussian Mixtures from Datasets Distributed by Features}

\aistatsauthor{ Pedro Valdeira \And Cl\'{a}udia Soares \And  Jo\~{a}o Xavier }

\aistatsaddress{ CMU\textsuperscript{$\dagger$}, IST\textsuperscript{$\ddagger$} \And  NOVA\textsuperscript{$\S$} \And IST\textsuperscript{$\ddagger$}} ]

\begin{abstract}
Expectation Maximization (EM) is the standard method to learn Gaussian mixtures.
Yet its classic, centralized form is often infeasible, due to privacy concerns and computational and communication bottlenecks.
Prior work dealt with data distributed by examples, horizontal partitioning, but we lack a counterpart for data scattered by features, an increasingly common scheme (e.g. user profiling with data from multiple entities). To fill this gap, we provide an EM-based algorithm to fit Gaussian mixtures to Vertically Partitioned data (VP-EM).
In federated learning setups, our algorithm matches the centralized EM fitting of Gaussian mixtures constrained to a subspace.
In arbitrary communication graphs, consensus averaging allows VP-EM to run on large peer-to-peer
networks as an EM approximation. This mismatch
comes from consensus error only, which vanishes exponentially fast with the number of consensus rounds.
We demonstrate VP-EM on various topologies
for both synthetic and real data, evaluating its approximation
of centralized EM and
seeing that it outperforms the available benchmark.
\end{abstract}

\section{INTRODUCTION} Expectation Maximization (EM) is both the standard method for density estimation of Gaussian Mixture Models (GMMs) and a popular clustering technique.
Yet these two key tasks do not constitute an exhaustive list of
EM applications.
\blfootnote{
	Email: \texttt{pvaldeira@cmu.edu}.\\
	$\dagger$Carnegie Mellon University\\
	$\ddagger$Instituto Superior T\'{e}cnico\\
	$\S$NOVA School of Science and Technology
}
From its formalization \citep{dempster77} up to today, EM has been widely employed, with applications ranging from semi-supervised learning \citep{ghahramani93} to topic modeling \citep{blei03}. EM remains an active research subject, as seen in the prolonged effort to study its convergence, from \citet{wu83} until today \citep{balakrishnan14, daskalakis17, kunstner21}.
\begin{figure}[t]
	\centering
	\includegraphics[width=1\columnwidth]{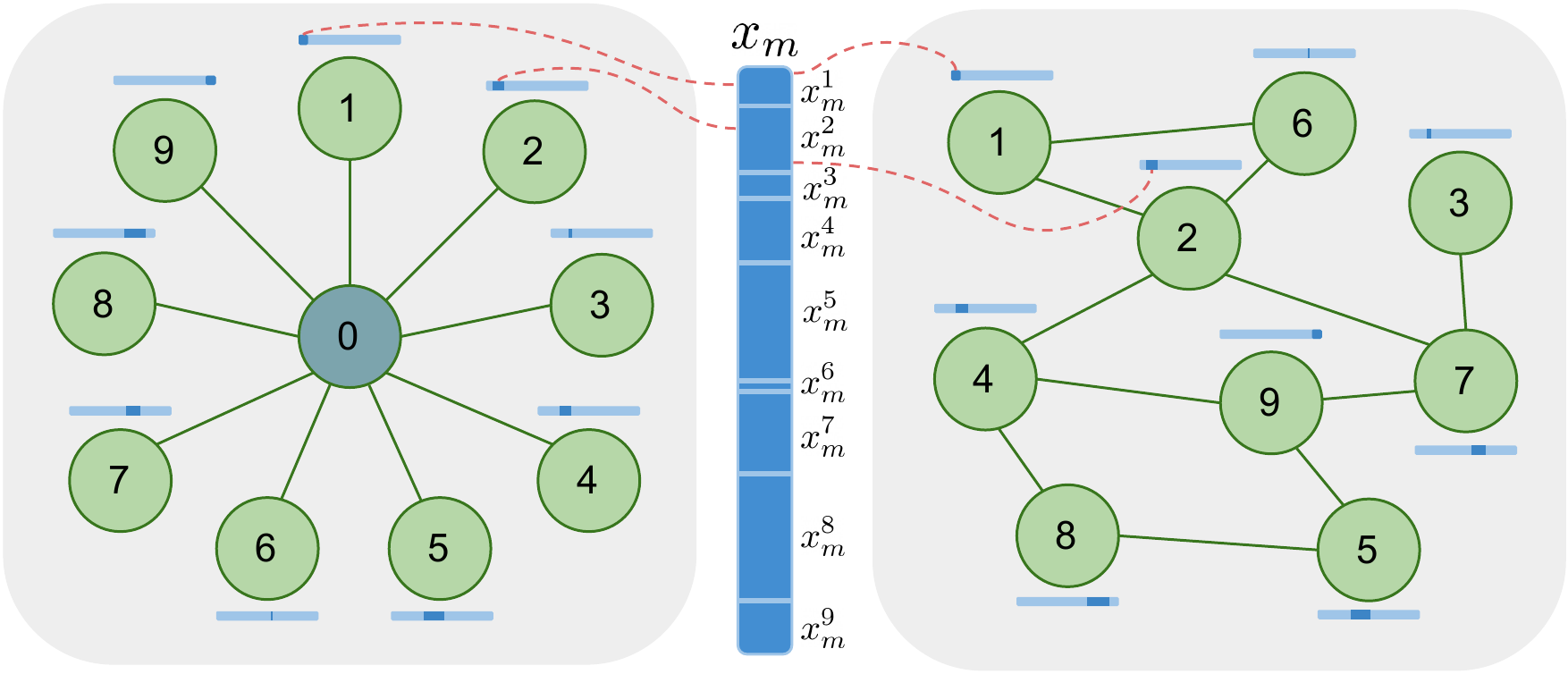}
	\caption{Each example $x_m$ is partitioned by features, either in federated learning (left) or fully decentralized (right) networks of agents. We want an unsupervised learning algorithm that allows the agents to perform density estimation and clustering in collaboration without requiring their data to be centralized.}
	\label{fig:distributed_diagram}
\end{figure}

Employing EM to fit GMMs, a textbook application, is a particularly popular unsupervised learning tool. This density estimation is used to model subpopulations and approximate densities---even mixtures of diagonal Gaussians, given enough components, can approximate
generic densities to an arbitrary error~\citep{sorenson71}---making Gaussian mixtures exceptionally expressive models. Further, GMMs allow for ellipsoidal clustering, making them more flexible than the spherical clusters of the popular $k$-means algorithm. Moreover,
while $k$-means performs hard clustering, GMMs allows for soft clustering,
expressing uncertainty associated with the clustering
and handling clusters that are not mutually exclusive.

However, data is often too large, privacy sensitive, or both, preventing the use of a data center where conventional, centralized methods can be employed. Such setups call for
distributed approaches where data is treated while partitioned. The partitioning may come from a pre-processing step to cope with the scale of data \citep{Mann2009}, but it may also be the original layout of data collected by multiple entities \citep{Nedic2010}.
In either case, distributed learning allows us to spread computational costs
which are too high for a single agent.
Further, without privacy guarantees, it is often impossible
for different entities to collaborate and learn from joint data.
Yet, even when their data cannot be routed to a single agent
(regardless of its size), exchanging a summarized,
non-invertible function of the data may be acceptable\footnote{Many distributed algorithms, including ours, do not provide privacy guarantees directly. Rather, they allow for privacy-preserving learning to take place when tools such as differential privacy \citep{dwork2008} and homomorphic encryption \citep{rivest1978} are employed on top of them.}.
Thus, avoiding centralization circumvents privacy concerns.
Distributed learning is also key in systems where high speed is paramount \citep{Cheng2012} and communication delays prevent centralized approaches.

Scenarios requiring distributed approaches include large networks, like edge computing applications and personalized models \citep{wang2019}, and networks with fewer agents, such as organizations (e.g. healthcare systems \citep{rieke2020}) with siloed data. 

One distributed learning scheme that has been drawing a remarkable amount of attention, both in academia and in industry, is Federated Learning (FL) \citep{mcmahan2017}, where multiple agents (clients) collaborate to learn a shared model under the coordination of a central server without sharing their local data. This scheme is characterized by a star communication graph and a server-client architecture (see Figure~\ref{fig:distributed_diagram}).
Nevertheless, FL has some limitations, such as having a single point of machine failure and being prone to communication bottlenecks \citep{lian2017} as pointed out in \citet{kairouz2021}.

The more general \emph{fully decentralized} learning circumvents these
issues by having no central parameter server (agents form a peer-to-peer network, see Figure~\ref{fig:distributed_diagram}).
Fields like multi-agent control and wireless sensor networks employ extensive work on optimization under fully decentralized schemes, such as \citet{Olshevsky2009, Nedic2009, Duchi2012, Qu2018}, and references therein. While these works are often formulated as generic optimization problems, rather than designed for a specific learning task, they tend to be motivated by applications where data is distributed by examples (horizontally
partitioned), as is made clear, for example, in \citet{Forero2010}. 
Closer to our work, multiple fully decentralized algorithms
use EM to fit GMMs in horizontally partitioned setups, such as \citet{Nowak2003, Kowalczyk2005, Gu2008, forero2008, bhaduri09, safarinejadian2010, weng2011, Altilio2019}. Related density estimation tasks have also been considered \citep{hu2007, hua2015, Dedecius2017}.

In these horizontally partitioned settings, loss
functions tend to naturally decouple across agents. Yet in many applications, such as online shopping records data and social network interactions, the \emph{features} are collected across different providers.  In these setting, or  when the number of examples, $M$, and their dimension, $d$, verify $d\gg M$, as
in the natural language processing and
bioinformatics applications in \citet{Boyd2011}, the data benefits from \textit{Vertical Partitioning} (VP) \citep{yang2019}.

The flexible fully decentralized VP learning, in specific, allows us to cope with the arbitrary communication graphs observed in IoT, teams of robots, and wireless sensor networks applications too. (E.g., agents deployed to learn the behavior of a random field across a region, such as the concentration of a pollutant.)
The literature for the VP setting is not as abundant as its horizontal counterpart.
\citet{Vaidya2003} tackle VP $k$-means clustering, focusing on privacy, rather than aiming for full decentralization; \citet{grozavu2010} perform VP clustering resorting to self-organizing maps; \citet{ding2016} addresses $k$-means clustering for FL; and \citet{Fagnani2013} deals with density estimation, assuming a restricted GMM where each agent is associated with a component of the mixture.
Note that these and our work consider unsupervised learning, yet VP supervised learning has also been studied \citep[e.g.][]{ying2018}. 
For more VP learning algorithms, see \citet{yang2019}.

We present an EM-based method for fitting Gaussian mixtures to VP data both in federated learning and fully decentralized setups (VP-EM), allowing for the scalability and privacy we set out to achieve. These setups are illustrated in Figure~\ref{fig:distributed_diagram}, where the agents may represent anything from companies to sensors.

Our main contributions are as follows:
\begin{itemize}[leftmargin=12pt, itemsep=0pt]
	\item Formulating the task of fitting GMMs in vertically partitioned schemes and identifying its challenges;
	\item Proposing a solution to these challenges by partitioning
	parameters and constraining their space;
	\item Framing the VP-EM algorithm for FL setups, with the same monotonicity guarantee as the classic EM;
	\item Extending VP-EM to peer-to-peer setups, coping with additional challenges and exploring the topology to offer a range of data sharing options;
	\item Equipping VP-EM with stopping criteria.
\end{itemize}

\section{EM AND GAUSSIAN MIXTURES}
EM performs (local) maximum likelihood estimation in models where the likelihood function for parameter $\theta$ depends on observed data $X$, but also on latent variables $Z$,
\begin{equation}
	\mathcal{L} (\theta) = \log p\left(X\,|\, \theta \right)
	= \log \sum_Z p\left(X,Z \,|\,  \theta \right).
	\label{eq:em_objective}
\end{equation}
For mixture models, $Z$ often indicates the mixture component which, if known, would make the likelihood function concave, with a closed-form solution. (For missing data, $Z$ corresponds to missing portions of the examples.) Yet, in reality, $Z$ is a random variable, thus maximizing the complete-data likelihood $p\left(X,Z \,|\,  \theta \right)$ is not a well-posed problem. However, if we fix $\theta$ to some $\theta^{t}$, we can infer $Z$ using its posterior $p\left(Z \,|\, X,\theta^{t}\right)$ and computing the expected complete-data likelihood (E-step), thus defining a surrogate objective,
\[\mathcal{Q}\left(\theta \,|\,  \theta^{t}\right)
= \mathbf{E}_{Z | X,\theta^{t}}\left[
\log p(X,Z\mid \theta)
\right].\]
With a well-defined objective, we update $\theta$ using standard optimization tools (M-step), solving
\[\theta^{t+1}=
\text{argmax}_\theta \, \mathcal{Q}\left(\theta \,|\,  \theta^{t}\right).\]
After initializing $\theta$, the EM iterates between E- and M-steps until some stopping criterion is met.

\paragraph{Theorem 1.} \emph{The EM is monotonically non-decreasing on its objective,}
\[\mathcal{L}\left( \theta^{t+1} \right) \geq \mathcal{L}\left( \theta^{t} \right).\]
This property, shown in \citet{dempster77}, often results in convergence of~\eqref{eq:em_objective} to a (local) maximizer.

A GMM with $K$ components, combined with weights $\left\{ \pi_k\right\}_{k = 1}^K$ on the probability simplex, has the density
\begin{equation}
	p_\theta(x ) = {\textstyle \sum\nolimits_{k = 1}^K} \pi_k\, {\mathcal N}\left( x ; \mu_k,
	\Sigma_k \right),
	\label{eq:ptheta}
\end{equation}
where $x,\mu_k \in {\mathbf R}^d$, $\Sigma_k \in {\mathbf R}^{d\times d}$, $\theta = \left\{ \pi_k, \mu_k, \Sigma_k\right\}_{k = 1}^K= \left\{ \pi, \mu, \Sigma\right\}$, and the Gaussian distribution is given by
\begin{equation}
	\label{eq:mvn}
	{\mathcal N}\left( x ; \mu, \Sigma \right) =
	\lvert 2 \pi \Sigma \rvert^{-\frac{1}{2}}
	\exp\left( -( x- \mu )^T \Sigma^{-1} ( x - \mu )/2\right).
\end{equation}

When using EM to fit a GMM to $\mathcal{X}=\{x_m\}_{m=1}^M\subseteq\mathbf{R}^d$, we have $\mathcal{Z}=\{z_m\}_{m=1}^M$, where $z_m\in\mathbf{R}^K$ indicates the component from which $x_m$ is sampled. The E-step boils down to updating the parameters of the posterior
\begin{equation}
	\label{eq:estep_gmm}
	{\gamma}_{mk}^{t+1} =
	\mathbf{E}_{Z | X,\theta^{t}}\left[ z_{mk} \right]
	\propto \tilde{\gamma}_{mk}^{t+1} = \pi_{k}^t {\mathcal N}\left( x_m ; \mu_{k}^t, \Sigma_{k}^t \right),
\end{equation}
where ${\gamma}_{mk}^t$ is obtained by normalizing $\tilde{\gamma}_{mk}^t$ over $k$. This step performs a soft assignment of each example $m$ to each cluster $k$ (notice ${\gamma}_{mk}^{t+1}\geq 0$ and $\mathbf{1}^T{\gamma}_{m}^{t+1}=1$). ${\gamma}_{mk}^t$ is called the \emph{responsibility} that component $k$ takes for $x_m$ at iteration $t$.

The M-step updates the GMM parameters as follows:
\begin{subequations}
	\label{eq:mstep_gmm}
	\begin{align}
		{\textstyle \pi_{k}^{t+1} =} &
		{\textstyle \frac{1}{M} \sum\nolimits_{m=1}^{M} \gamma_{mk}^{t+1}}
		\label{eq:mstep_pi},\\
		{\textstyle \mu_{k}^{t+1} =} &
		{\textstyle \frac{1}{M\pi_{k}^{t+1}} \sum\nolimits_{m=1}^{M} \gamma_{mk}^{t+1}x_m}
		\label{eq:mstep_mu},\\
		{\textstyle \Sigma_{k}^{t+1} =} &
		{\textstyle \frac{1}{M\pi_{k}^{t+1}} \sum\nolimits_{m=1}^{M} \gamma_{mk}^{t+1}
			\left(x_m - \mu_{k}^{t+1} \right) \left(x_m - \mu_{k}^{t+1} \right)^T}.
		\label{eq:mstep_sigma}
	\end{align}
\end{subequations}
A more detailed analysis of EM and its application to GMM can be found, for example, in \citet{bishop2006}.

\section{DISTRIBUTED AVERAGING CONSENSUS}
\label{sec:consensus}
Let $\psi \in {\mathbf R^N}$ be scattered across a set of $N$ agents in an undirected communication graph $\mathcal{G}= \left( {\mathcal V}, {\mathcal E} \right)$, where $\mathcal{V}=\{n\}_{n=1}^N$ is the vertex set and and $\mathcal{E}\subseteq \mathcal{V}^2$ is the edge set, with $\{n,n^\prime\}\in\mathcal{E}$ if and only if $n$ and $n^\prime$ are connected in the graph. Entry $\psi_n $ is held by agent~$n$.

Distributed averaging consensus, or simply consensus \citep{saber2003,xiao2004}, delivers an approximate average to all agents, $\langle \psi \rangle =  \mathbf{1}^T\psi/ N$. This is achieved by an iterative update of the state of each agent, initialized as $z_n^0 = \psi_n$, by combining it linearly with the states of its neighbors,
\begin{equation} z^{s+1} = W z^{s}, \label{eq:cons} \end{equation}
where $z^{s} = (z_1^{s}, \dots, z_N^{s} ) \in {\mathbf R}^N$ is  the state of the network. The sparsity pattern of weight matrix $W$ matches the underlying communication graph, that is, $\{n, n^\prime \} \not\in {\mathcal E} \implies W_{nn^\prime}=0$, thus ensuring the consensus update~\eqref{eq:cons} requires local communications only (i.e., between neighbors). Further, we need $W$ to be a symmetric matrix with eigenpair $(1,\mathbf{1})$ and all other eigenvectors (which are not proportional to ${\mathbf 1}$) associated with eigenvalues whose absolute value is strictly less than 1. These properties are secured, for example, by having $W$ be the Metropolis weights matrix, or by taking $W=I - \alpha L$, where $L$ is the Laplacian of the graph and $\alpha\in \left( 0, \lambda_{\max}(L) \right)$ is a constant. For more details on Metropolis weights and graph spectral theory see, e.g., \cite{chung1994,xiao2005}.  For such $W$, repeated application of~\eqref{eq:cons} brings the state of
all agents to the average of the distributed data, that is,
$\lim_{s \rightarrow +\infty} z^s = \langle \psi \rangle {\mathbf 1}.$ Further, although exact agreement of the agents on
$\langle \psi \rangle$ is only guaranteed as the number of iterations
$s$ approaches infinity, consensus converges
exponentially fast. Thus, in practice, we can stop consensus after a finite number of iterations, say $S$,
with $z_n^S \approx \langle \psi \rangle$ at each agent
$n$.

\section{CHALLENGES OF LEARNING GMMS FROM DATA DISTRIBUTED BY FEATURES}
As in Figure~\ref{fig:distributed_diagram}, let $\mathcal{X}=\{x_m\}_{m=1}^M$ be a dataset where each example $x_m$ is distributed across $N$ agents connected by a undirected communication graph $\mathcal{G}$,
\begin{equation} \label{eq:xm_partitioning}
	x_m=
	(x_{m}^1, \dots, x_{m}^n, \dots, x_{m}^N)
	\in \mathbf{R}^d,
\end{equation}
where agent $n$ holds partition $x_{m}^n\in \mathbf{R}^{d_n}$ for all $m$. One challenge of such partitioning is that, in contrast to the horizontal counterpart, the data is generally not iid across agents. This, in turn, is often reflected in loss functions that do not naturally decouple across agents.

If we naively run the EM on such setup, multiple problems come up, the first being the issue of storing the model parameters $\theta = \left\{ \pi_k, \mu_k, \Sigma_k\right\}_{k = 1}^K$ and the responsibilities $\gamma\in \mathbf{R}^{M\times K}$. Each entry $\gamma_{mk}$ is associated with an entire data point $m$ and mixture component $k$,
having no feature-associated dimension that scatters naturally in our setting, thus each agent holds all of $\gamma$. Since we seek scalability and privacy along features, not examples (although, as mentioned later, a batch VP-EM removes dependencies on $M$), this is not a problem. In contrast, $\theta$ has feature-specific information, thus storing all of $\theta$ in each agent would lead to memory requirements proportional to $d$ on the agents, namely through $\mu_k\in\mathbf{R}^d$ and (worse) $\Sigma_k\in\mathbf{R}^{d\times d}$. Although memory costs are often not the main bottleneck, we want to avoid this dependency, mainly due to its implications for computational and communication costs, which we now see by examining an EM iteration.

Let us see how to partition $\theta^t$ across the network.
From \eqref{eq:estep_gmm} and \eqref{eq:mvn}, we get, with a slight abuse of notation in the equality (we drop the $2\pi$ term, which cancels out when normalizing), that the E-step requires computing
\begin{equation} \label{eq:log_prop_gamma}
	\log \tilde{\gamma}_{mk}^{t+1} =
	\log \pi_{k}^t
	-\frac{1}{2} \log \lvert \Sigma_{k}^t \rvert
	-\frac{1}{2} \left\lVert x_m- \mu_{k}^t \right\rVert_{(\Sigma_{k}^t)^{-1}}^{2},
\end{equation}
where $\left\lVert v \right\rVert_{A}^{2}=v^TAv$. Note that $\log \pi_{k}^t$ is innocuous, as it does not depend on $d$. Further, $x_m- \mu_{k}^t$ does
depend on $d$, but it can computed distributedly by partitioning $\mu_k$ such that each agent $n$ holding $x_{m}^n$ stores the corresponding $\mu_{k}^n$ for all $k$.
The troublemaker is $\Sigma_{k}^t$.
Notice how all computations involving this term (determinant, inverse, and quadratic product) hinder a distributed E-step, coupling entries that respect features observed at arbitrary nodes in the network. To deal with these challenges, on top of partitioning $\Sigma_{k}^t$, we see we may need to approximate it, decoupling the
determinant and inverse into smaller terms
of size independent from $d$ and preventing $(\Sigma_{k}^t)^{-1}$ from coupling arbitrary entries of $x_m- \mu_{k}^t$ in the quadratic product.

In the M-step, we see from \eqref{eq:mstep_pi} that we can locally update $\pi_k$ in all agents, using the local $\gamma$, thus all agents keep $\pi$, as the EM iterates. Likewise, \eqref{eq:mstep_mu} confirms that the aforementioned partitioning of $\mu_{k}$ is viable, since $n$ holds all the terms required to update $\mu_{k}^n$ with
\begin{equation} \label{eq:dist_mstep_mu}
	{\textstyle \left( \sum\nolimits_{m=1}^{M} \gamma_{mk}^{t+1}(x_m)_n \right) / \left(M\pi_{k}^{t+1}\right) }.
\end{equation}
In \eqref{eq:mstep_sigma}, $\Sigma_k$ raises another problem. We see its partitioning and approximation must decouple the outer product. Imposing some sparsity on $\Sigma_k$ emerges as an appealing approach, since entries assumed to be zero need not be estimated. We will see that an appropriate sparsity can indeed decouple the outer product.

We conclude that, to avoid communications between arbitrary agents in the network, we must allocate $\mu$ and $\Sigma$ taking into consideration where in the graph each feature is observed. For $\mu$, we saw that a simple partitioning, similar to that of the data, suffices for the EM to iterate distributedly. The following sections will detail how to partition $\Sigma$.

\section{VP-EM FOR FEDERATED LEARNING}
\label{sec:vpem_fl}
\begin{figure}[t]
	\centering
	\includegraphics[width=\columnwidth]{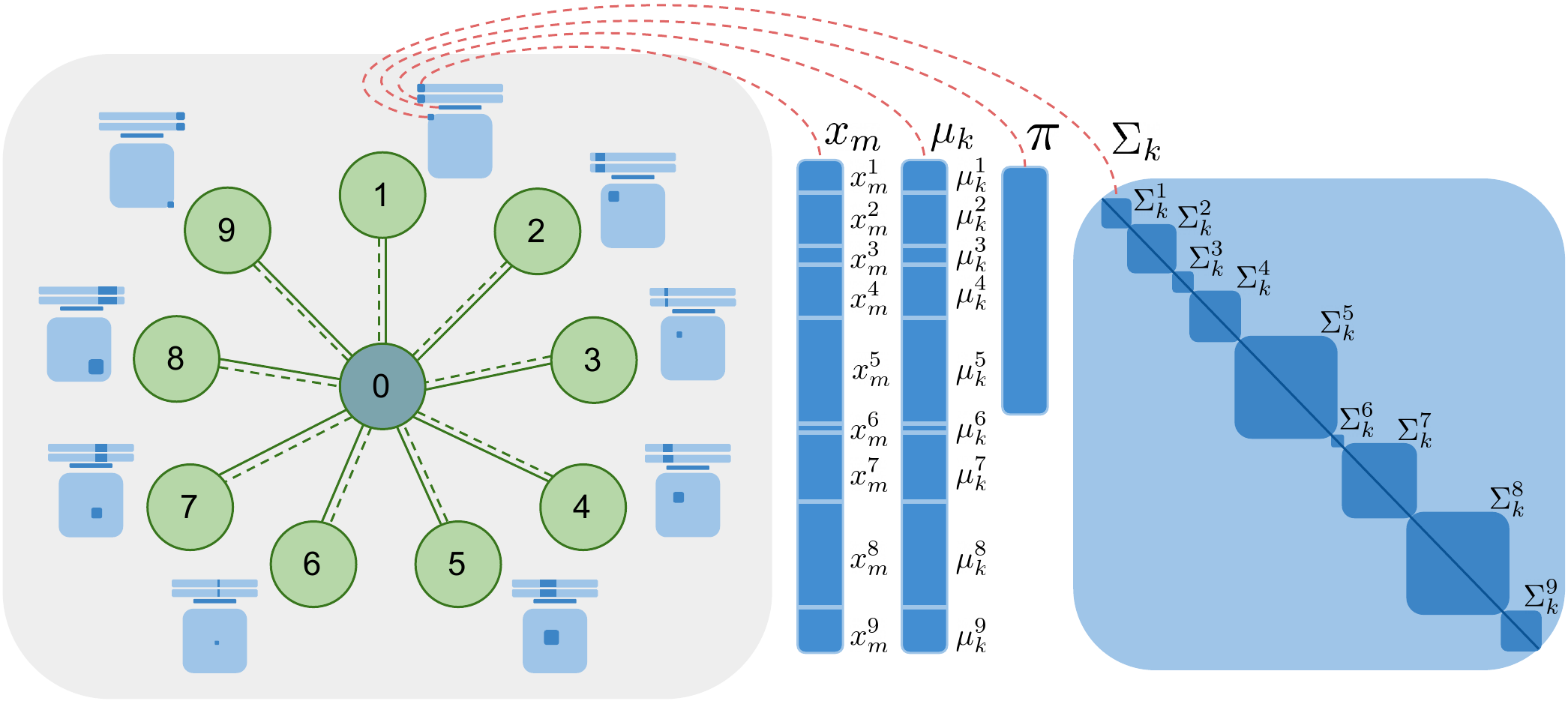}
	\caption{The VP-EM for federated learning. The E-step requires communications between clients and server (solid line), while the M-step requires local client computations only (dashed line): we alternate between a local step and a connected one.
		Data and model parameters are partitioned, with the darker blue representing the portions at each client.
		Since $\gamma$ is neither, we omit it,
		but it is held in its entirety at every client.
		All $m$ and $k$ have the same partitioning.}
	\label{fig:fl_emsteps}
\end{figure}
Let $\mathcal{X}$ be scattered over a network of data centers, as in \eqref{eq:xm_partitioning}. The \emph{server}, node~$0$, is connected to all \emph{clients}, nodes $n=1,\dots,{N}$, which have no direct communication channels between them, that is, $\mathcal{E}=\left\{
\left\{ 0,1\right\}, \left\{ 0,2\right\},\dots,\left\{ 0,N\right\} \right\}$.
Each client $n$ holds copies of $\gamma$ and $\pi$, as well as
the partition
$\mu_{k}^n\in \mathbf{R}^{d_n}$ for all $k$.
(We omit the EM iteration superscript to avoid clutter.)
In FL, we overcome the challenges associated with $\Sigma_k$
by constraining it to be a block-diagonal matrix where each block $\Sigma_k^n\in\mathbf{R}^{d_n\times d_n}$ matches
the features of $x_{m}^n$. Entry $(i,j)$
of $\Sigma_k$, $[\Sigma_k]_{ij}$, is assumed zero when features $i$ and $j$ are observed by different clients, as seen in Figure~\ref{fig:fl_emsteps}.
\begin{algorithm}[tb]
	\caption{VP-EM for Federated Learning}
	\label{alg:vpem_fl}
	\textbf{Input}: $X$\\
	\textbf{Parameter}: $K$\\
	\textbf{Output}: ${\textstyle \hat{\theta}=\left\{ \hat{\pi}_k, \hat{\mu}_k, \hat{\Sigma}_k\right\},\,\hat{\gamma}}$
	\begin{algorithmic}[1] 
		\State Choose $\theta^0 = \left\{ \pi^0, \mu^0, \Sigma^0\right\}$ \Comment{Same $\pi^0$ across agents}
		\For{$t=0,1,\dots$}
		\For{$m\in[M]$, $k\in[K]$} \Comment{E-step (update $\gamma$)}
		\State Each client $n$ locally computes its term
		\Statex \hskip\algorithmicindent \hskip\algorithmicindent
		in~\eqref{eq:estep_sum} and sends it to the server;
		\State Server adds the $N$ terms and returns \eqref{eq:estep_sum} \Statex \hskip\algorithmicindent \hskip\algorithmicindent to the clients;
		\State Each client computes \eqref{eq:log_prop_gamma} locally and
		\Statex \hskip\algorithmicindent \hskip\algorithmicindent normalizes over $k$, arriving at \eqref{eq:estep_gmm};
		\EndFor
		\State Each client $n$ computes 
		\eqref{eq:mstep_pi}, \eqref{eq:dist_mstep_mu}, and \eqref{eq:dist_mstep_sigma},
		\Statex \hskip\algorithmicindent updating $\theta_n$;
		\Comment{M-step ($\theta^t\gets\theta^{t+1}$)}
		\If {converged}
		\State \textbf{break}
		\EndIf
		\EndFor
		\State \textbf{return} $\theta^t$, $\gamma$
	\end{algorithmic}
\end{algorithm}
Note that $[\Sigma_k]_{ij}=0$ does not mean features $i$ and $j$
are independent, as we are dealing with GMMs. (That would
be true if $K=1$.)
This structure simplifies the inversion of $\Sigma_k$ to a block-wise inversion and its determinant to a product of determinants of the blocks,
$\lvert \Sigma_k \rvert = 
\prod_{n=1}^{N}\lvert \Sigma_k^n \rvert$, decoupling $\log\lvert \Sigma_k \rvert$ as a sum over agents. A similar decoupling over $n$ is obtained in the quadratic term in \eqref{eq:log_prop_gamma}. Thus, in the E-step, the server performs a single sum,
\begin{equation} \label{eq:estep_sum}
	{\textstyle
	\sum_{n=1}^{N}\log\lvert \Sigma_k^n \rvert + \left\lVert x_m^n-
	\mu_{k}^n \right\rVert_{\left(\Sigma_{k}^n\right)^{-1}}^{2},
	}
\end{equation}
which is then sent to the clients who, in turn, multiply it by $-{1}/{2}$, add their local copy of $\log \pi_{k}$, and normalize over $k$ locally, arriving at $\gamma_{mk}$.

From \eqref{eq:mstep_sigma}, since only non-zero entries must be estimated, we see the outer product now decouples, allowing each agent $n$ to update $\Sigma_{k}^n$ locally with
\begin{equation} \label{eq:dist_mstep_sigma}
	{\textstyle \frac{1}{M\pi_{k}} \sum\nolimits_{m=1}^{M} \gamma_{mk}
		\left(x_m^n- \mu_{k}^n\right) \left(x_m^n - \mu_{k}^n \right)^T.}
\end{equation}

Algorithm \ref{alg:vpem_fl} outlines VP-EM for federated learning, where $[a]=\{1,\dots,a\}$ and $\theta_n$ are the parameters stored by agent $n$, $\{\pi_k,\mu_k^n, \Sigma_k^n\}_{k=1}^K $. Note that, as depicted in Figure~\ref{fig:fl_emsteps}, only the E-step requires communications and that each data partition remains in the agent that observed it.
For implementation purposes, we propose tracking the
log-likelihood, $\mathcal{LL}$, using its
diminishing increment as a stopping criterion
, as is often done in classic EM.
$\mathcal{LL}$ is obtained as a byproduct of the E-step,
requiring no additional communications
(see supplementary material for further details).

\paragraph{Corollary 1.} \emph{The VP-EM algorithm for federated learning is monotonically non-decreasing on its objective,}
\[\mathcal{L}\left( \theta^{t+1} \right) \geq \mathcal{L}\left( \theta^{t} \right).\]
This follows directly from Theorem 1, given that VP-EM matches the EM when constrained to a subspace.

Importantly, we can drop the dependency on $M$
of the computational complexity of each EM iteration.
We do this by updating only a subset of the entries of $\gamma$ in
the E-step, which, in turn, allows for an online update of the
parameters in the M-step, since only the associated subset
of the terms in the sums in \eqref{eq:mstep_pi}, \eqref{eq:mstep_mu}, and \eqref{eq:mstep_sigma} changes.
This batch version of VP-EM follows directly from the application of the incremental version of the EM \citep{neal93} and is explained in greater detail in the supplementary material.

\section{VP-EM FOR FULLY DECENTRALIZED LEARNING}
Let $\mathcal{X}$ be (again) scattered across a communication graph, as in \eqref{eq:xm_partitioning}. We now handle arbitrary graph topologies, assuming only that $\mathcal{G}$ is connected.
This section can be seen as
an extension of the VP-EM for FL, as it based on similar assumptions. Yet now we
\textbf{(1)} cope with fully decentralized sums;
\textbf{(2)} allow for less sparsity to be imposed on $\Sigma_k$
when privacy concerns permit.
To achieve the latter point 
while still attaining a block-diagonal structure,
we exploit the flexibility
of the communication graphs describing fully decentralized setups.
In particular,
we explore the fact that
the richer topologies allow for
informative notions of locality,
based on the distance (number of hops)
to the agent observing each feature.
This allows us to think
in terms of a degree of privacy.
(In FL, all clients were a hop
away from the server, so moving the data even
one hop away
from the clients observing it leads to centralization
in the server.)

\begin{figure}[t]
	\centering
	\includegraphics[width=0.95\columnwidth]{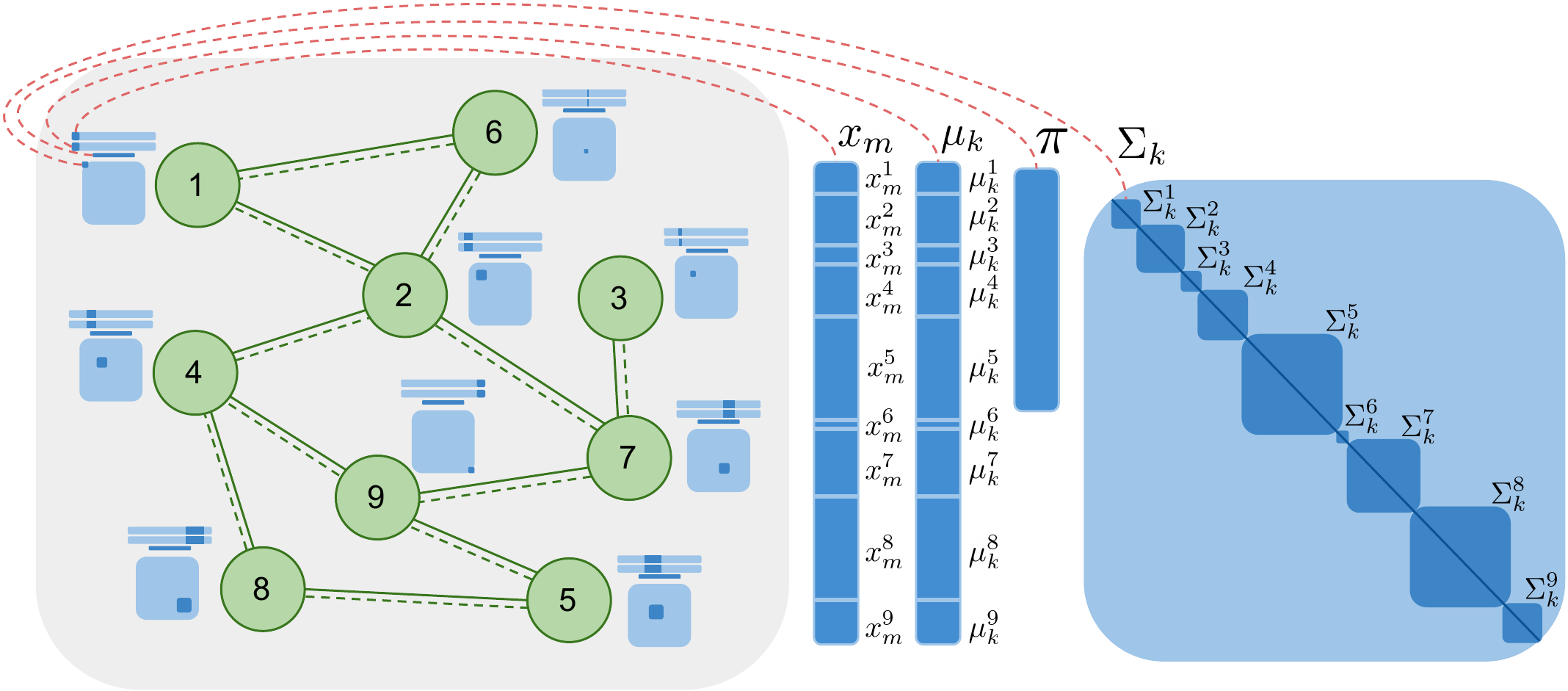}
	\caption{
		The VP-EM for fully decentralized learning. The E-step requires local communications (solid line), while the M-step does not
		(dashed line): we alternate between a local step
		and a connected one.
		Data and model parameters are partitioned, with the darker blue representing the portions at each agent.
		Since $\gamma$ is neither, we omit it,
		but it is held in its entirety at each agent.
		All $m$ and $k$ follow the same partitioning.
		We represent $\Sigma_k$ for the
		$0$-hop scheme to avoid clutter in the illustration of $\Sigma_k$.
		A diagram contrasting
		the sparsity of $\Sigma_k$ for $0$-hop and $1$-hop schemes (less sparse) can be found in the supplementary material.
	}
	\label{fig:p2p_emsteps}
\end{figure}

But how do we choose the new block-diagonal structure of $\Sigma_k$?
First, note we can still employ the
sparsity constraint used in FL
($[\Sigma_k]_{ij}\neq0$
when features $i$ and $j$ are observed by the same agent).
In fact, this is our only option
when privacy concerns forbid any data sharing.
We call this a $0$\textit{-hop} scheme,
as it allows for the communication of features
to agents within $0$ hops of the one observing them,
$\{\}$. (See Figure~\ref{fig:p2p_emsteps}.)
Yet, when some local data sharing is allowed,
we should explore this, leveraging
the richer graph topologies to define a notion of distance
to the node observing each feature, which can be used
to estimate more entries $\Sigma_k$ while keeping a
block-diagonal structure.
Graph clustering \citep{Schaeffer2007} proves useful.
We propose a simple procedure
to choose the blocks of $\Sigma_k$,
which runs only once,
before the EM-based iterations.

\paragraph{Graph Clustering.}
We want to choose the largest blocks possible for $\Sigma_k$,
to constrain our search space the least possible,
while respecting a degree of privacy requirement
expressed by $h$ when opting for an $h$-hop communication scheme.
The graph clustering algorithm takes as input the
graph $\mathcal{G}$ and outputs
the sparsity pattern of a (block-diagonal) matrix.
For $1$-hop data sharing,
the $b$th cluster (or \textit{hub}), $\mathcal{H}_b$, corresponds
to the union of some node $n$ with its neighbors $\mathcal{N}_n$, $\mathcal{H}_b=n\cup \mathcal{N}_n$. In general, $h$-hop hubs include
all nodes up to $h$ hops away from the \emph{root} node $n$.
The algorithm is greedy, picking
the largest $h$-hop hub in the graph and extracting it
(settling ties at random).
For $1$-hop hubs, this means picking
the node with the highest degree (number of neighbors).
We find and extract the first hub, $\mathcal{H}_1$,
such that
$\text{root}(\mathcal{H}_1)=\argmax_{n\in\mathcal{V}}\lvert
\mathcal{N}_n\rvert$.
Let $p_1=\lvert\mathcal{H}_1\rvert$,
nodes in $\mathcal{H}_1$ are assigned
numbers $1, 2, \ldots, p_1$ in some order.
We repeat the process for the residual graph,
$\text{root}(\mathcal{H}_2)=\argmax_{n\in{\mathcal V} - {\mathcal H}_1}\lvert
\mathcal{N}_n\rvert$,  where $\mathcal{N}_n$ are
now the neighbors in ${\mathcal V} - {\mathcal H}_1$, extracting ${\mathcal H}_2$.
Let $p_2=\lvert\mathcal{H}_2\rvert$,
nodes in ${\mathcal H}_2$ are assigned
numbers $p_1+1, \ldots, p_1+p_2$.
We repeat the process until the residual graph is void. For $1$-hop schemes, each hub is a star-shaped graph:
a node stands at the root and its neighbors,
the \textit{leaf} nodes, surround it.
We assume $[\Sigma_k]_{ij}=0$
when features $i$ and $j$ are
observed by nodes in different hubs,
obtaining a block-diagonal
$\Sigma_k$ with blocks $\left\{
\Sigma_{k}^b
\right\}_{b=1}^B$ for all $k$,
where $B$ is the number of hubs (and thus blocks) obtained.
This procedure can be run in a distributed manner,
by finding the maximum degree of the network via flooding.
\begin{algorithm}[tb]
	\caption{Fully Decentralized VP-EM}
	\label{alg:vpem_p2p}
	\textbf{Input}: $X$\\
	\textbf{Parameter}: $K$\\
	\textbf{Output}: ${\textstyle \hat{\theta}=\left\{ \hat{\pi}_k, \hat{\mu}_k, \hat{\Sigma}_k\right\},\,\hat{\gamma}}$
	\begin{algorithmic}[1] 
		\State Graph clustering
		\State Data sent from leaves to roots
		\State Choose $\theta^0 = \left\{ \pi^0, \mu^0, \Sigma^0\right\}$ \Comment{Same $\pi^0$ across roots}
		\For{$t=0,1,\dots$}
		\For{$m\in[M]$, $k\in[K]$} \Comment{E-step (update $\gamma$)}
		\State The root of each hub $b$ computes its term
		\Statex \hskip\algorithmicindent \hskip\algorithmicindent
		in~\eqref{eq:estep_sum_p2p} and sends it to the leaves of
		$b$;
		\State All agents engage in consensus averaging;
		\State Each root performs local computations
		\Statex \hskip\algorithmicindent \hskip\algorithmicindent  to arrive at a local estimate of \eqref{eq:estep_gmm};
		\EndFor
		\State Each root updates $\theta_b$, computing
		\eqref{eq:mstep_pi} and hub
		\Statex \hskip\algorithmicindent
		versions of \eqref{eq:dist_mstep_mu} and \eqref{eq:dist_mstep_sigma};
		\Comment{M-step (update $\theta$)}
		\If {converged}
		\State \textbf{break}
		\EndIf
		\EndFor
		\State \textbf{return} $\theta^t$, $\gamma^t$
	\end{algorithmic}
\end{algorithm}

As mentioned, the $h$-hop scheme allows us
to take advantage of data sharing protocols,
increasing collaboration while respecting privacy constraints.
(Note that, in FL setups, $h$-hop with $h>0$
results in a centralized approach.)
Yet, beyond privacy,
$h$ can also be seen as parameterizing
the trade-off between a more expressive model,
where VP-EM approximates the centralized EM
more closely, and a parsimonious scheme,
where we focus on reducing the costs at each agent
(which increase, for root nodes, as $h$ increases).

Having defined the sparsity of $\Sigma_k$
and assigned some agents the role of root nodes,
let us understand the memory requirements.
Note that these requirements are just
a generalization of the ones for the FL scheme,
which, despite not requiring a graph clustering algorithm,
can be seen as taking every node to be a hub (i.e.
$\mathcal{H}_1=\{1\},\dots,\mathcal{H}_N=\{N\}$ and thus $B=N$).
The memory requirements that follow
apply to any $h$-hop.

VP-EM requires the root$(\mathcal{H}_b)$
to hold: the features observed by all nodes in
$\mathcal{H}_b$, $x_m^b$, for all $m$ (the
data is sent from the leaves to the root at the start of the algorithm);
the corresponding features of $\mu_{k}$, $\mu_{k}^b$,
for all $k$;
the $b$th block $\Sigma_{k}$, $\Sigma_{k}^b$, for all $k$; parameters $\pi_k$, for all $k$;
and $\gamma_{mk}$ for all $k$ and $m$.
(Note that there are no memory requirements on the leaf nodes.)

In the E-step, our approach is similar to that of \eqref{eq:estep_sum},
but now the sum decouples as follows:
\begin{equation} \label{eq:estep_sum_p2p}
	\sum_{b=1}^{B}\underbrace{\log\lvert \Sigma_k^b \rvert + \left\lVert x_m^b- \mu_{k}^b \right\rVert_{\left(\Sigma_k^b\right)^{-1}}^{2}}_{\mathcal{Q}_{mk}^b}.
\end{equation}
We obtain this sum by \textbf{(1)} computing
$\mathcal{Q}_{mk}^b$ at the root of $\mathcal{H}_b$,
which then sends it to any leaf agents of $\mathcal{H}_b$ and
\textbf{(2)} engaging all agents in a consensus
where the states of the agents in hub $\mathcal{H}_b$ are
initialized with $N\mathcal{Q}_{mk}^b/ | \mathcal{H}_b |$.
(We now have the result of \eqref{eq:estep_sum_p2p}
in every agent, up to consensus error.)
At this point, each root agent
multiplies the result of the sum by $-{1}/{2}$,
adds its local estimate of $\log \pi_{k}$,
and normalizes over $k$ locally, arriving at
its local estimate of $\gamma_{mk}$.

In the M-step, for each hub $\mathcal{H}_b$,
its root node updates its
local estimate of $\pi_{k}$,
as in \eqref{eq:mstep_pi}, using its
local estimate of $\gamma_{mk}$, while
$\mu_k^b$ and $\Sigma_k^b$ are updated
similarly to \eqref{eq:dist_mstep_mu} and
\eqref{eq:dist_mstep_sigma},
but hub partitions $x_m^b$, $\mu_k^b$, and $\Sigma_k^b$
replace agent partitions $x_m^n$, $\mu_{k}^n$, and $\Sigma_{k}^n$. Also, $\pi_{k}^{t+1}$ and $\gamma_{mk}$ are now local estimates of the root node of $\mathcal{H}_b$. These estimates
can differ across hubs, but go to the same value as we increase the number of consensus rounds.
In fact, if an infinite number of rounds were possible,
Corollary~1 would hold for this setting too.

VP-EM for general topologies is outlined in Algorithm~\ref{alg:vpem_p2p}, where
$\theta_b$ are the parameters stored by the root of hub $b$, $\{\pi_k,\mu_k^b, \Sigma_k^b\}_{k=1}^K $. It is
important to note a characteristic of our algorithm:
the existence of a distributed stopping criterion.
Stopping criteria are a matter of practical concern when
implementing such iterative algorithms, yet
it is notoriously difficult to find one in
fully decentralized setups
(especially without additional communications).
VP-EM enjoys a natural distributed stopping criterion.
Agents can track the global likelihood function
without additional communications.
We use its diminishing increment as a stopping criterion,
as is often done in classic EM.
(For further details, see supplementary material.)

As in FL, the dependency on $M$ can be dropped by implementing a batch VP-EM that follows directly from \citet{neal93}.
In fact, the importance of the batch version
is even greater for general topologies,
since, even if more EM iterations are needed before $\mathcal{LL}$ plateaus,
each iteration has significantly lower
communication costs.
(Using full-batch, we run the consensus algorithm
$M\times K$ times per iteration to compute $\gamma$.)

\section{EXPERIMENTS}
We demonstrate our algorithm for density estimation and clustering,
comparing it with centralized EM in both,
to evaluate our approximation, and with the
clustering benchmark. (We are not aware of
any algorithm performing a similar density estimation
in our setup).
We test VP-EM
in various distributed settings
of interest, resorting to synthetic and real data,
focusing on fully decentralized schemes,
since VP-EM for FL behaves
as the centralized EM
fitting of a block-diagonal GMMs, a well-studied topic.
Throughout the experiments, we use 100 rounds of consensus for each average,
approximating an infinite number rounds.

\subsection{Density Estimation}
\begin{figure}[t]
	\begin{subfigure}{.5\columnwidth}
		\centering
		\includegraphics[width=.93\linewidth]{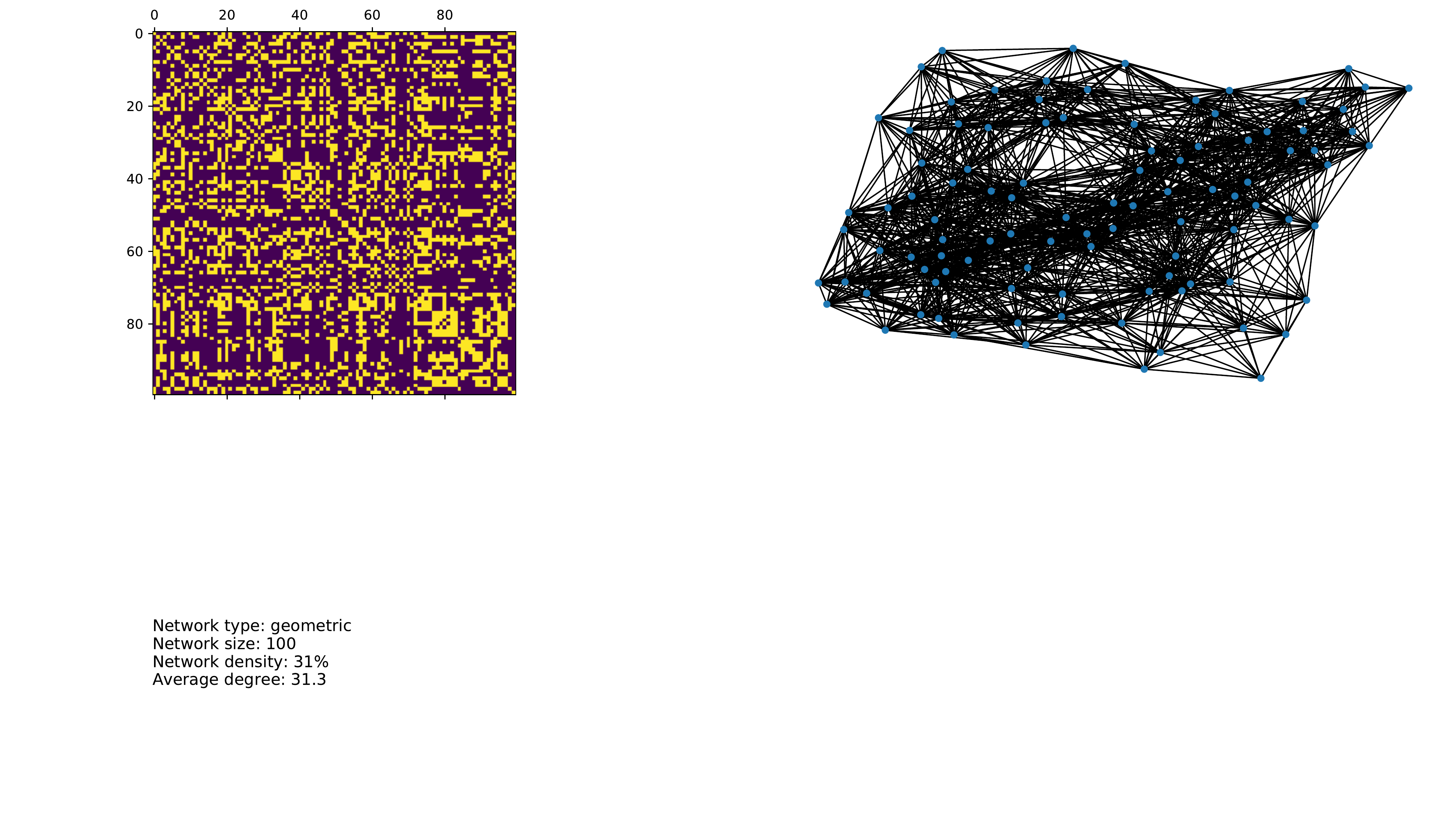}
	\end{subfigure}%
	\hfill
	\begin{subfigure}{.5\columnwidth}
		\centering
		\includegraphics[width=.87\linewidth]{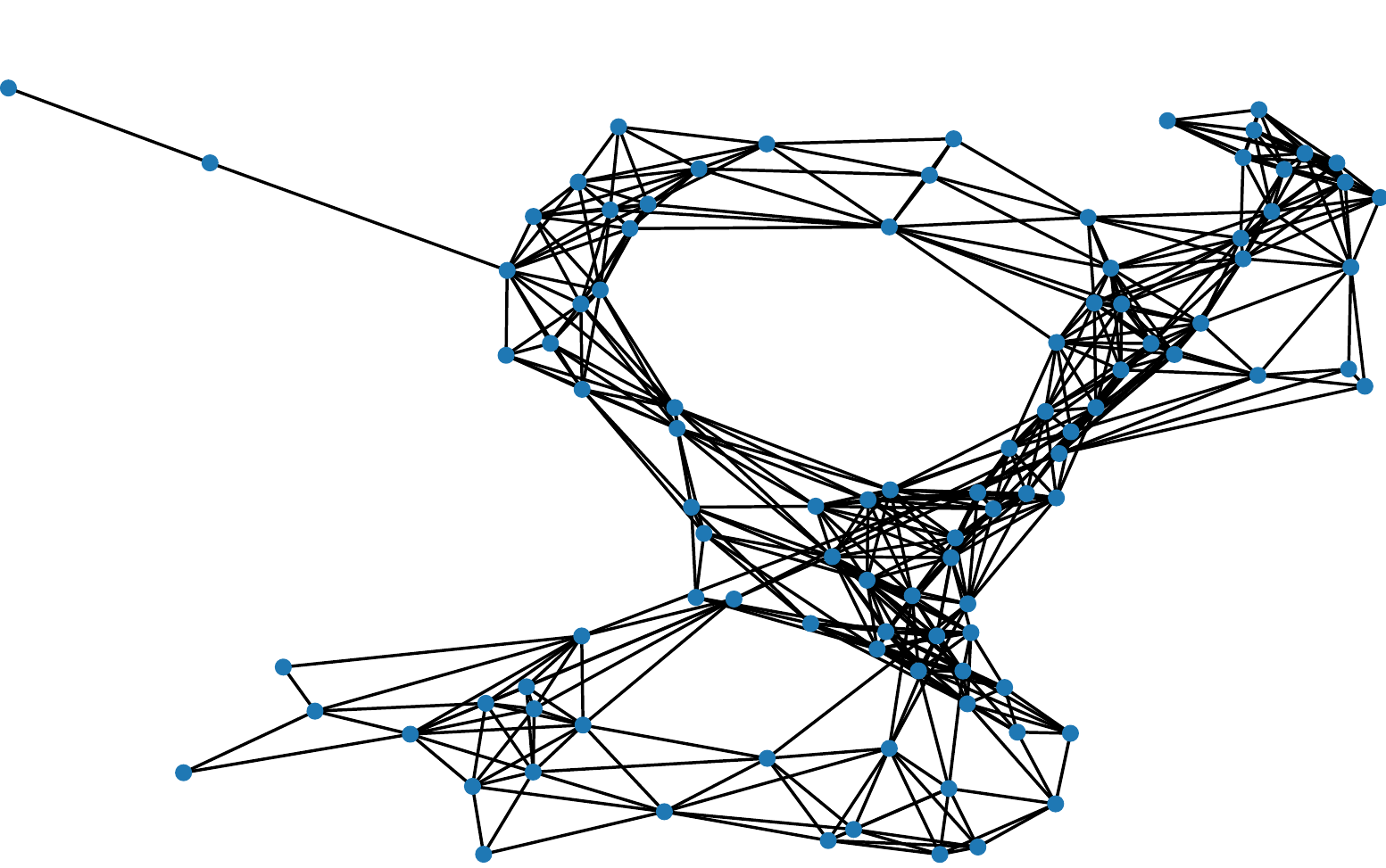}
	\end{subfigure}
	\caption{Random geometric networks used in
		synthetic data experiments, both with $N=100$:
		a dense network with an average degree of $31$ (left)
		and a sparse one with an average degree of $9.7$ (right).}
	\label{fig:fig_nets}
\end{figure}
\begin{figure}[t]
	\begin{subfigure}{.45\columnwidth}
		\centering
		\includegraphics[width=\linewidth]{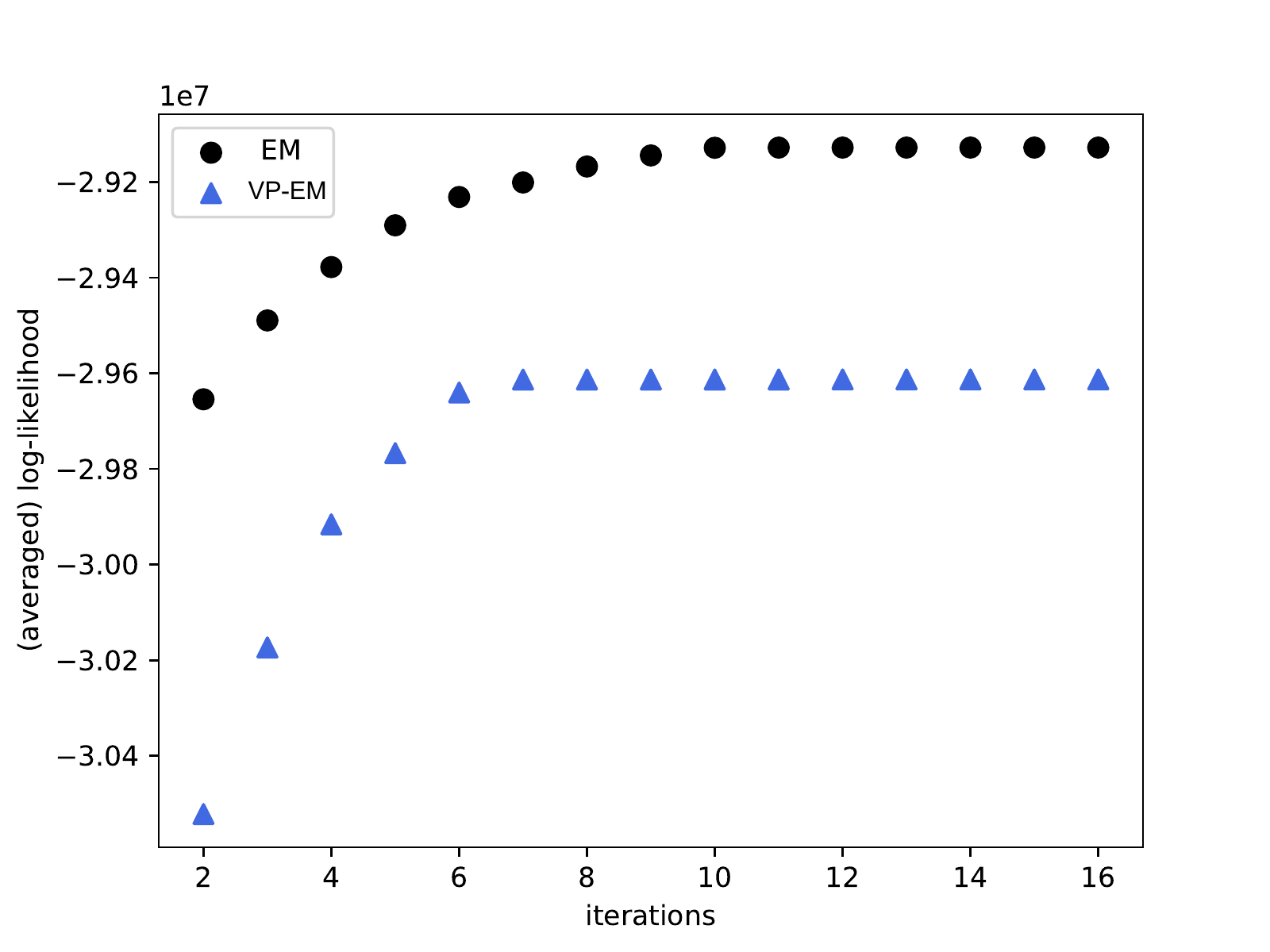}
	\end{subfigure}%
	\hfill
	\begin{subfigure}{.45\columnwidth}
		\centering
		\includegraphics[width=\linewidth]{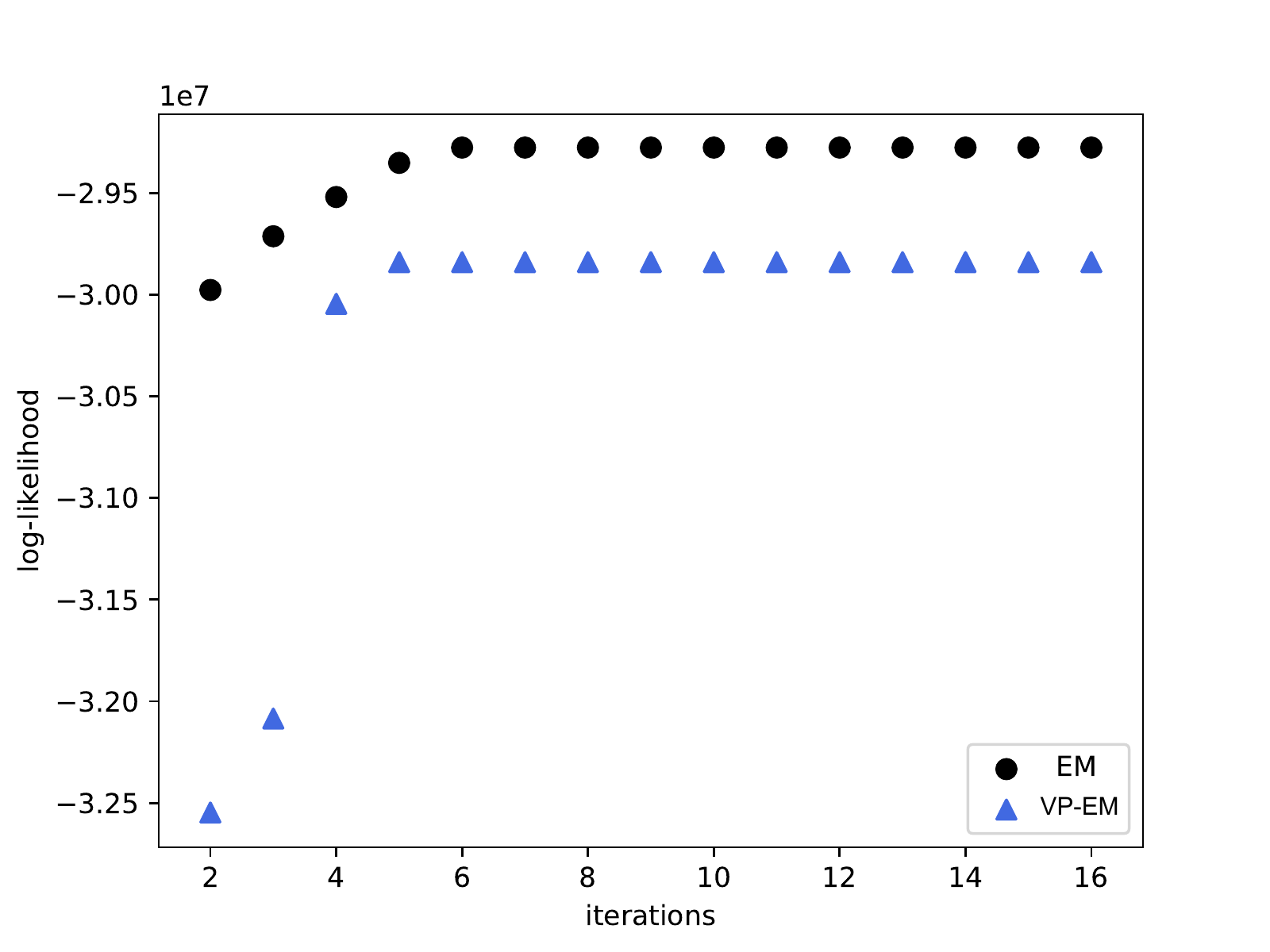}
	\end{subfigure}
	\caption{Log-likelihood trajectories for the synthetic data experiments: averaged per iteration over 5 runs on
		dense graphs (left); and run on sparse graph (right).
	}
	\label{fig:fig_liks}
\end{figure}
\paragraph{Synthetic data.}
We illustrate the scalability of VP-EM on five random
geometric networks with $N = 100$,
each fed $M = 150,000$ examples
sampled from a ground truth GMM with $K = 3$ components
created at random.
The five setups differ in the network instance,
data, and initializations, yet these networks
have the same $N$ and similar density of
connections (ratio between the
number of network links and the number of links in a complete
network, $\lvert\mathcal{E}\rvert/(N(N-1)/2)$). On the left sides of Figure~\ref{fig:fig_nets} and Figure~\ref{fig:fig_liks}, respectively, we have one of
the 5 networks and the evolution of $\mathcal{LL}$ averaged over the five runs, for both (centralized) EM and VP-EM. 
The simulations show $\mathcal{LL}$ increasing monotonically
before plateauing,
for both EM and our algorithm.
VP-EM plateaus at a lower value, as expected, due to using block-diagonal matrices $\Sigma_k$ which make our algorithm less expressive than the centralized EM,
which allows for fully dense matrices $\Sigma_k$. 



\begin{figure}[t]
	\begin{subfigure}{.45\columnwidth}
		\centering
		\includegraphics[width=.98\linewidth]{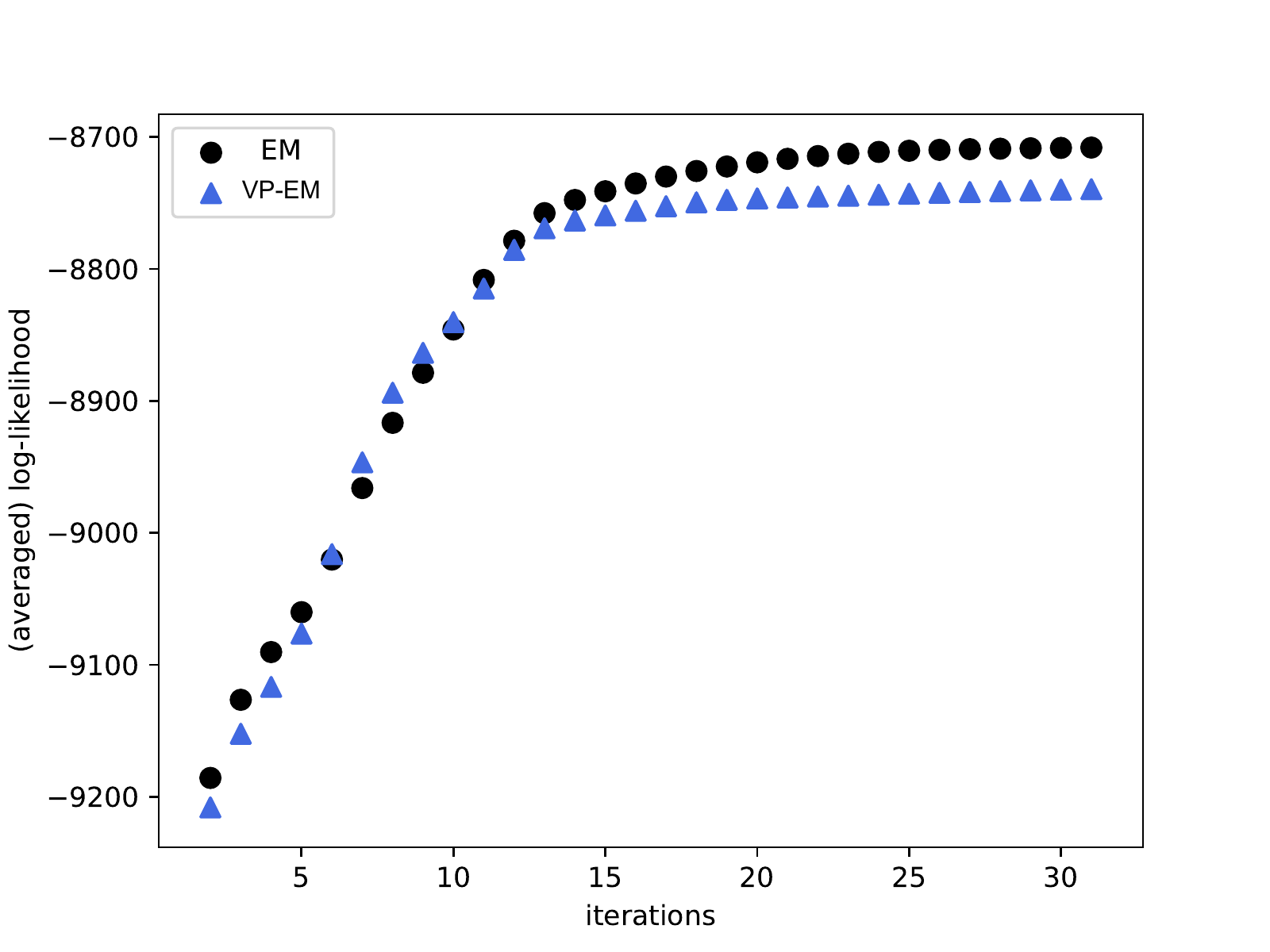}
	\end{subfigure}%
	\begin{subfigure}{.55\columnwidth}
		\centering
		\includegraphics[width=.98\linewidth]{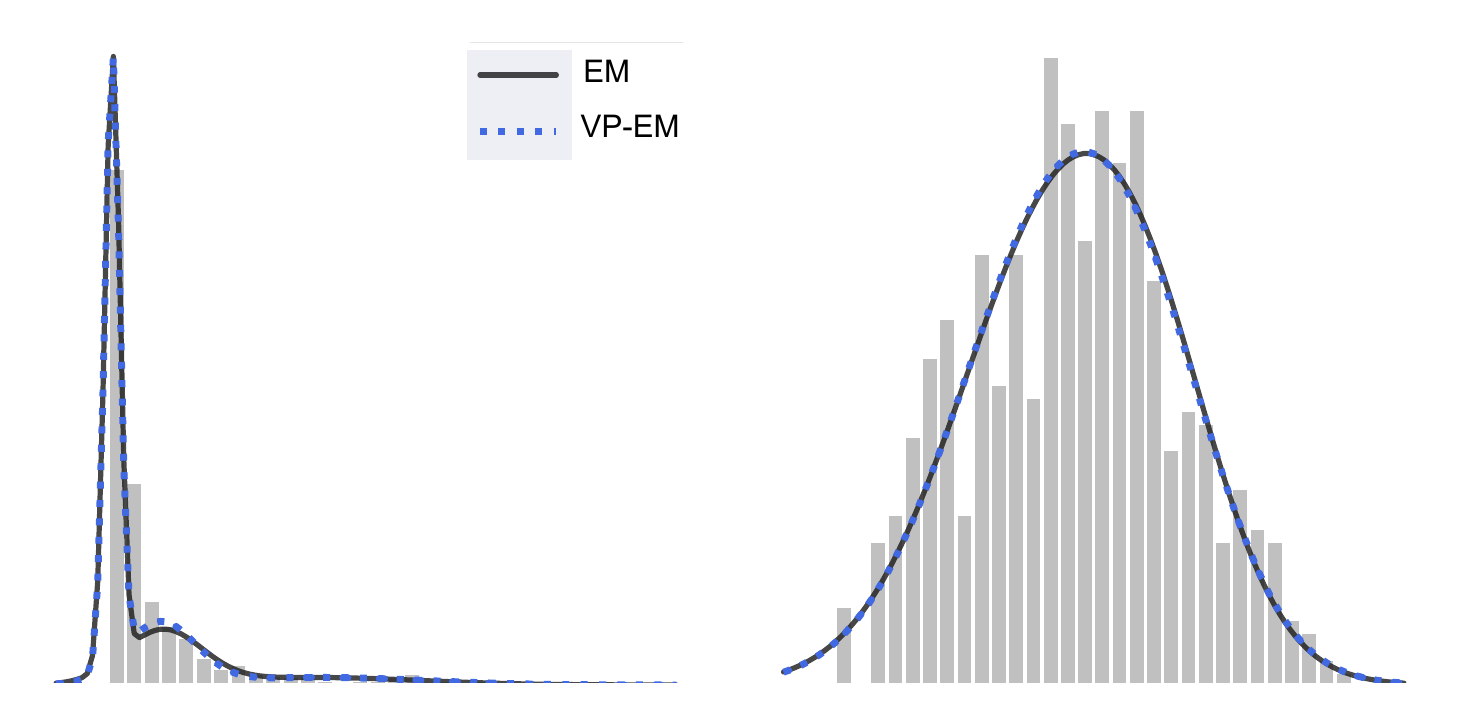}
	\end{subfigure}
	\caption{
		Log-likelihood trajectories for the real data experiment
		averaged per iteration over 5 runs (left).
		Marginal distributions of the EM and VP-EM GMMs
		superimposed on histograms of two features (right).
	}
	\label{fig:real_data_exp}
\end{figure}
In Figure~\ref{fig:fig_nets}, on the right, we present
the network used in an additional run of the algorithm,
with the same $K$, $N$, and $T$ as before,
but running VP-EM on a sparser network.
As seen in Figure~\ref{fig:fig_liks}, on the right, we
again observe a gap between the likelihoods attained by our algorithm and EM, due to the block-diagonal restriction
on $\Sigma_k$ by the former. Both $\mathcal{LL}_{\text{EM}}$ and $\mathcal{LL}_{\text{VP-EM}}$ remain monotonic increasing.

\paragraph{Real data.}
We now test VP-EM on real data
from a European public service queue system.
The data consists of a pool of $T_{\text{pool}} = 2, 650$
examples, each with $N=5$ features: daytime, weekday, queue,
number of people waiting, and the power transform \citep{yeo2000}
of
the moving average waiting times
for the relevant queue. 
We sample this pool five times at random to form five datasets with $T = 630$.
Each such dataset is then split by
features across a cycle of $N = 5$ agents, $\mathcal{E}=\left\{ \{1,2\}, \{2, 3\},\{3, 4\}, \{4,5\}, \{5,1\}
\right\}$. We chose a cycle graph due to it being one of the most challenging
topologies for information spreading. A different cycle graph can be seen
in Figure~\ref{fig:graphs}.

We ran (centralized) EM and VP-EM for each of
the five samples. The trajectory of
$\mathcal{LL}$ (averaged over the five
runs) is shown in Figure~\ref{fig:real_data_exp}. Again,
both algorithms converge monotonically and, again,
the gap between the two asymptotic values
of $\mathcal{LL}$ is  due to the sparsity imposed on $\Sigma_k$ by VP-EM.
(We ran a centralized EM with the
block-diagonal structure and saw its $\mathcal{LL}$ trajectory match that of 
VP-EM. Omitted to avoid clutter.)
Figure~\ref{fig:real_data_exp} also shows the GMMs fitted by the
centralized EM and VP-EM
superimposed on the histograms for two of the
features. The approximation seems good,
yet, when approximating
densities with GMMs, increasing $K$ may
narrow the gap between $\mathcal{LL}_{\text{VP-EM}}$ plateau and $\mathcal{LL}_{\text{EM}}$.
Here, a meager $K=3$ components allowed for good results. Yet, apart from a poor convergence
(which we can mitigate by starting the algorithm
multiple times with different initializations
and taking the best result), increasing $K$ should 
not lower $\mathcal{LL}$ and may in fact increase it.



\subsection{Clustering}
\begin{figure}[t]
	\begin{subfigure}{.28\columnwidth}
		\centering
		\includegraphics[width=\linewidth]{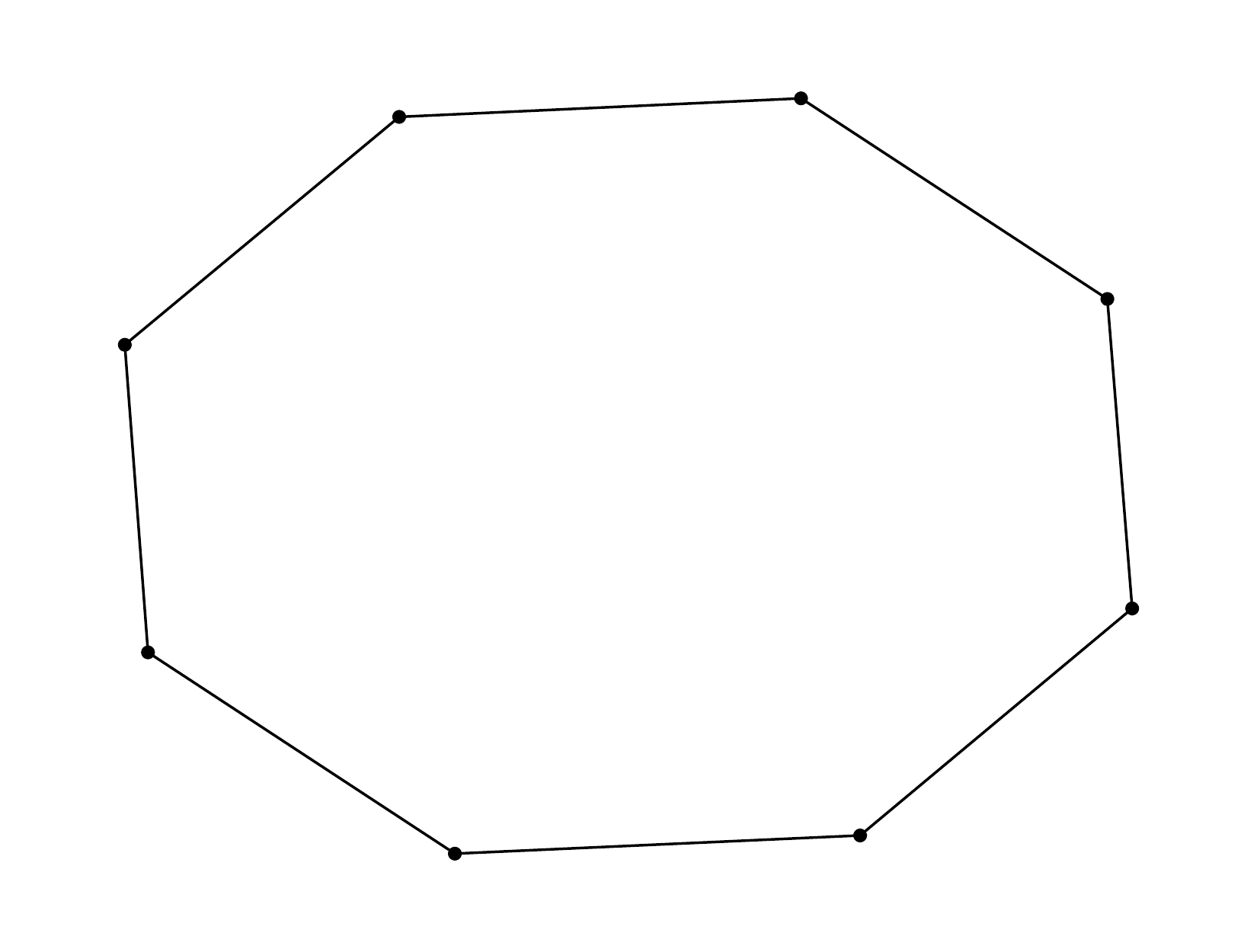}
	\end{subfigure}%
	\hfill
	\begin{subfigure}{.28\columnwidth}
		\centering
		\includegraphics[width=\linewidth]{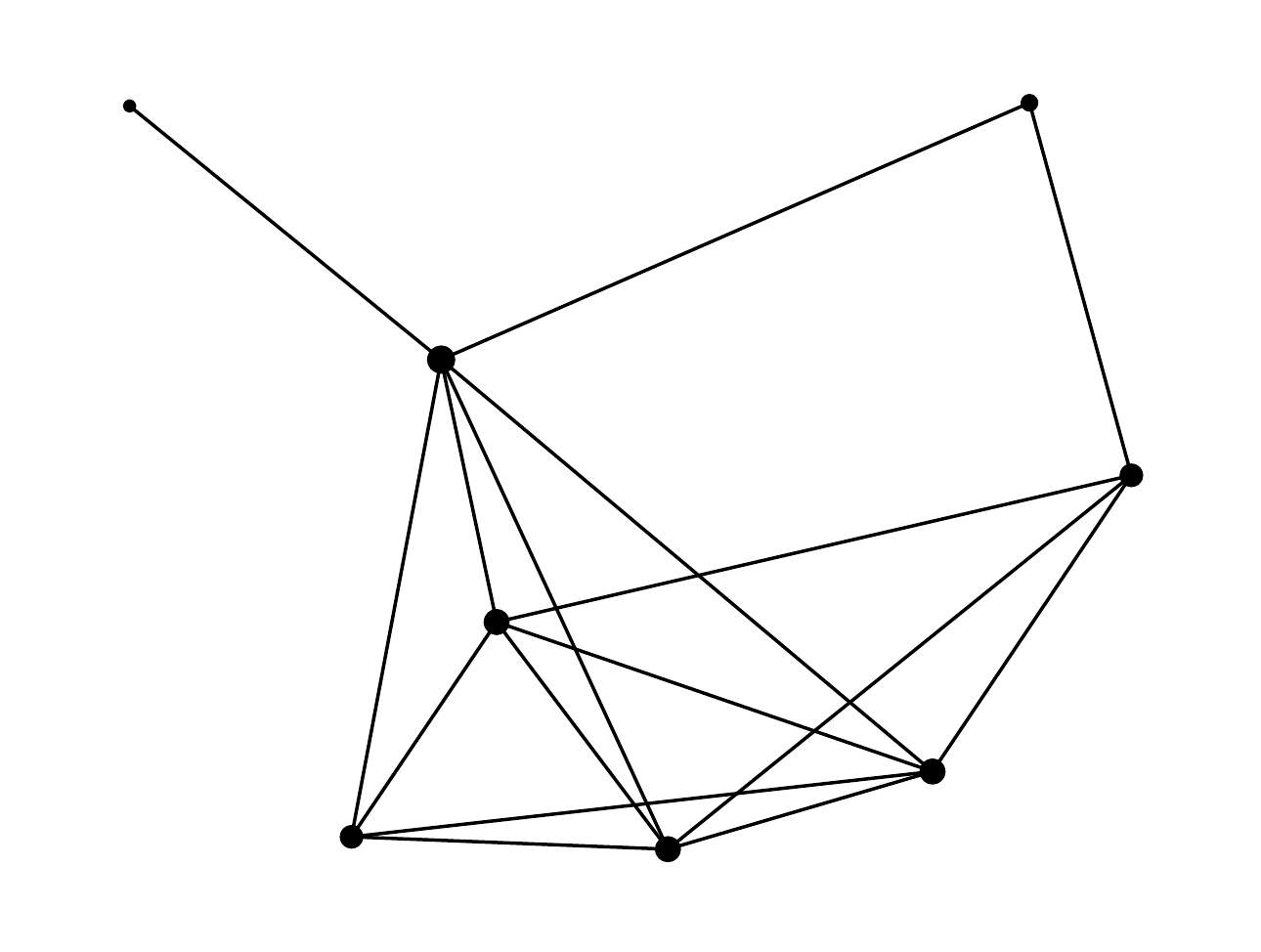}
	\end{subfigure}%
	\hfill
	\begin{subfigure}{.28\columnwidth}
		\centering
		\includegraphics[width=\linewidth]{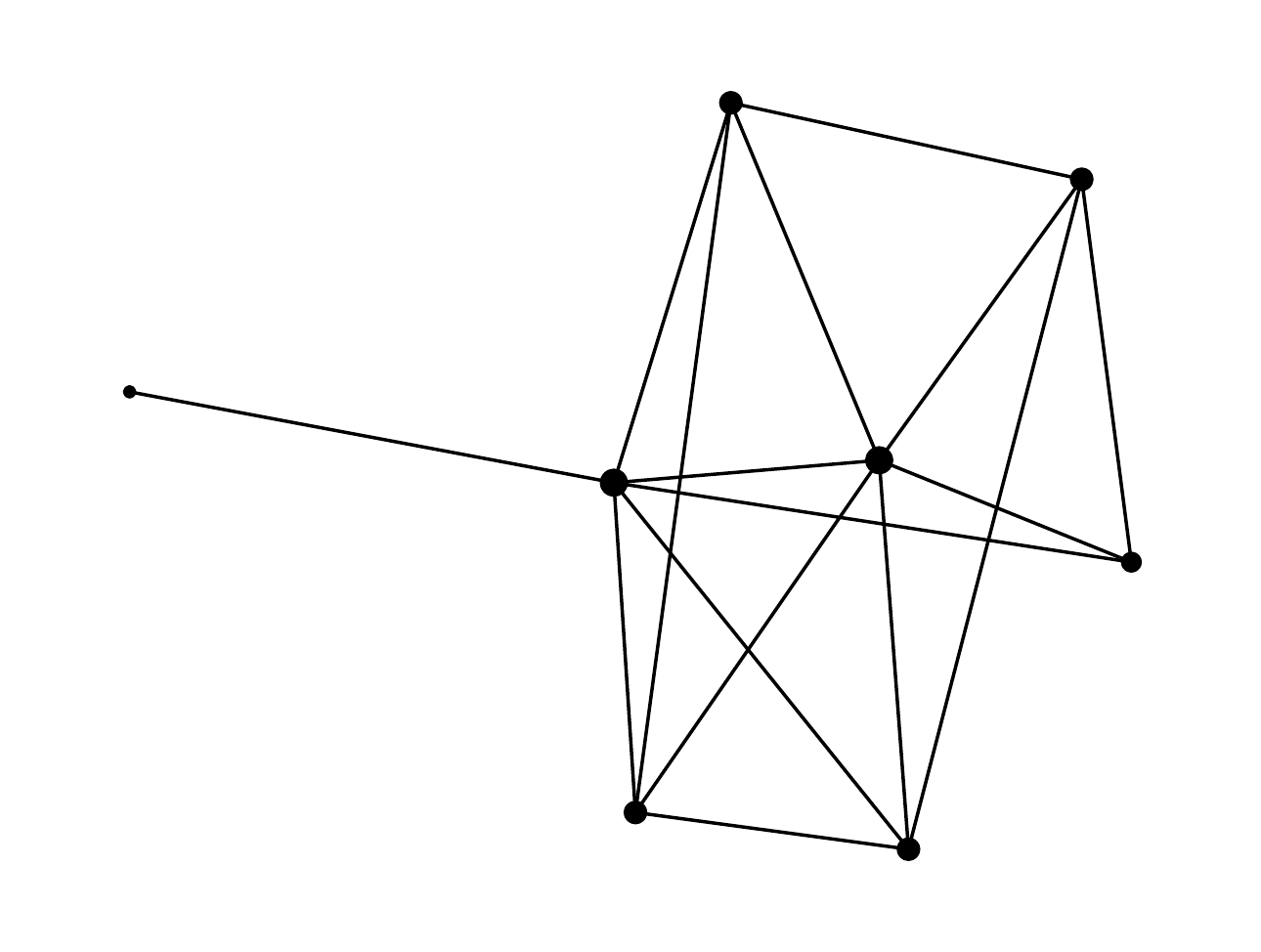}
	\end{subfigure}
	\caption{Cycle graph of VP$_{1}$ (left),
		geometric graph of VP$_{2}$ (middle), and
		scale-free graph of VP$_{3}$ (right). VP$_{2}$ and VP$_{3}$ have average degree 4.
	}
	\label{fig:graphs}
\end{figure}

\paragraph{Real data.}
For clustering, we do have a benchmark.
\citet{ding2016} performs $k$-means clustering in FL settings.
Thus, we now drop synthetic data altogether
to avoid biasing the results with the choice of ground truth, focusing on real data.
More precisely, we resort to data concerning
pulsar candidates
(measurements which may or not correspond to pulsars).
Pulsar discovery is an important task in astronomy and,
since
modern approaches produce millions of candidates,
a multitude of machine learning approaches has
emergenced  \citep[e.g.][]{Morello2014, Lyon2016}. 
We use the labeled data provided by \citet{Lyon2016} for clustering,
containing $M=17,898$ examples with $d=8$ features and
the correct label.
To facilitate running
the experiments, we upper bound the performance of the distributed $k$-means clustering with the standard, centralized $k$-means.
Since the dataset used was originally centralized,
it has no communication graph associated with it,
so we resort to synthetic networks.
(While VP-EM can run on fully decentralized setups,
if we were to run the benchmark,
instead of the centralized $k$-means,
it would require an FL scheme.)
More precisely, we run VP-EM on
three networks, as presented in Figure~\ref{fig:graphs}:
a cycle graph (setup VP$_{1}$), a geometric graph (setup VP$_{2}$), and a scale-free graph (setup VP$_{3}$).
In Table~\ref{tab:clustering_acc}, we see that
the (centralized) $k$-means is outperformed both
by EM and by VP-EM, in all setups. We actually even outperform the centralized EM, which may be due to overfitting by the centralized EM, which imposes no sparsity on $\Sigma_k$.
\begin{table}[tb]
	\caption{Clustering Accuracy} \label{tab:clustering_acc}
	\begin{center}
		\begin{tabular}{llllll}
			&$k$-means & EM & VP$_{1}$ & VP$_{2}$ & VP$_{3}$  \\
			\hline
			Acc.(\%)&76.7&84.3&85.7&84.4&87.2 \\
		\end{tabular}
	\end{center}
\end{table}

\section{Discussion}
We formulate the task of fitting GMMs to data distributed
by features and identify its challenges.
We propose an algorithm that copes with these
challenges, both in FL and in fully decentralized setups,
by partitioning parameters and constraining their space.
As VP-EM generalizes from FL to fully decentralized schemes,
we lose the monotonicity guarantee of the classic EM,
but get an opportunity to exploit flexible graph topologies,
allowing for a range of data sharing options which can relax the constraint on the parameter space.
We demonstrate VP-EM on fully decentralized setups,
using a large number of consensus rounds.
In the future, it would be interesting to study the algorithm in communication-critical setups.
Our results show that, in addition to outperforming the
benchmark, VP-EM approximates well the
centralized EM for both density estimation and clustering.

\bibliography{aistats}

\end{document}


%

%

\onecolumn
\aistatstitle{Decentralized EM to Learn Gaussian Mixtures from Datasets Distributed by Features:
Supplementary Materials}

\section{COMPUTING THE LOG-LIKELIHOOD}
We use the fully decentralized learning notation in this section,
since the federated learning scheme can be seen as a particular case
of this setting, which differs only in that it allows for the computation
of exact averages.

We saw that, in the E-step, we want to compute
\[
\log \tilde{\gamma}_{mk} =
\log \pi_{k}
-\frac{1}{2}
\left(\log \lvert \Sigma_{k} \rvert
+ \left\lVert x_m- \mu_{k} \right\rVert_{(\Sigma_{k})^{-1}}^{2}\right)
,
\]
where we omit the EM iteration superscripts to avoid clutter.
In the distributed case, the two terms being multiplied by
$-\frac{1}{2}$ can be decomposed as
\[
\sum_{b=1}^{B}\underbrace{\log\lvert \Sigma_k^b \rvert + \left\lVert x_m^b- \mu_{k}^b \right\rVert_{\left(\Sigma_k^b\right)^{-1}}^{2}}_{\mathcal{Q}_{mk}^b}.
\]
We obtain this in every agent (up to consensus error)
by \textbf{(1)} computing
$\mathcal{Q}_{mk}^b$ at the root of $\mathcal{H}_b$,
which then sends it to any leaf agents of $\mathcal{H}_b$ and
\textbf{(2)} engaging all agents in a consensus
where the states of the agents in hub $\mathcal{H}_b$ are
initialized with $N\mathcal{Q}_{mk}^b/ | \mathcal{H}_b |$.
At this point, each root agent
multiplies the result of the sum by $-{1}/{2}$,
adds its local estimate of $\log \pi_{k}$,
and normalizes over $k$ locally, arriving at
its local estimate of $\gamma_{mk}$:
\[
\label{eq:q_sum}
\log \tilde{\gamma}_{mk} =
\log \pi_{k}
-\frac{1}{2}
\sum_{b=1}^{B}\mathcal{Q}_{mk}^b, \quad \text{for all }
k\in[K],\,m\in[M],
\]
Note that
$\log \tilde{\gamma}_{mk}$
only differs from
$\pi_{k} {\mathcal N}\left( x_m ; \mu_{k}, \Sigma_{k} \right)$
in that it lacks the constant $-\frac{N}{2}\log (2\pi)$, which can be known by every agent.
Since each agent must store all $\gamma_{mk}$ and given that the log-likelihood corresponds to
\[ \mathcal{LL}(\theta)=\sum_{m=1}^{M}\log\left\{\sum_{k=1}^{K}\pi_k N(x_m\mid \mu_{k},\Sigma_{k})\right\}= \sum_{m=1}^{M}\log\left\{\sum_{k=1}^{K}(2\pi)^{-N/2}
\tilde{\gamma}_{mk}\right\},\]
it follows that $\mathcal{LL}(\theta)$ can be computed as a by-product of the E-step. More precisely, by multiplying $\tilde{\gamma}_{mk}$ by the constant $-\frac{N}{2}\log (2\pi)$, summing over all components in the mixture, taking the logarithm of that sum, and, finally, adding all these logarithms associated to each of the $M$ examples.

All this computations are performed \emph{locally}. That is, in order to compute the log-likelihood, providing us with a stopping criterion, we need not perform any additional communications.

\section{BATCH VP-EM}

The batch version of the VP-EM allows us to drop
the dependency on $M$
of the computational complexity of each EM iteration.
We do this by initializing both the parameters $\theta^0$
and the responsibilities $\gamma^0$ and then, in each EM iteration,
updating the subset of the entries of $\gamma$
associated with the batch (a subset of the data)
in the E-step, which, in turn, allows for an online update of the
parameters in the M-step, since only the associated subset
of the terms in the sums in the M-step updates changes.
This batch version of VP-EM follows directly from the
application of the incremental version of the EM \citep{neal93}.

As mentioned in the paper, the importance of the batch version
is even greater for general topologies,
since, even if more EM iterations are needed before $\mathcal{LL}$ plateaus,
each iteration has significantly lower
communication costs.
(Using full-batch, we run the consensus algorithm
$M\times K$ times per iteration to compute $\gamma$.)

\newpage

\section{THE IMPACT OF ALLOWING FOR MORE DATA SHARING}
\begin{figure}[h]
	\centering
	\includegraphics[width=0.75\columnwidth]{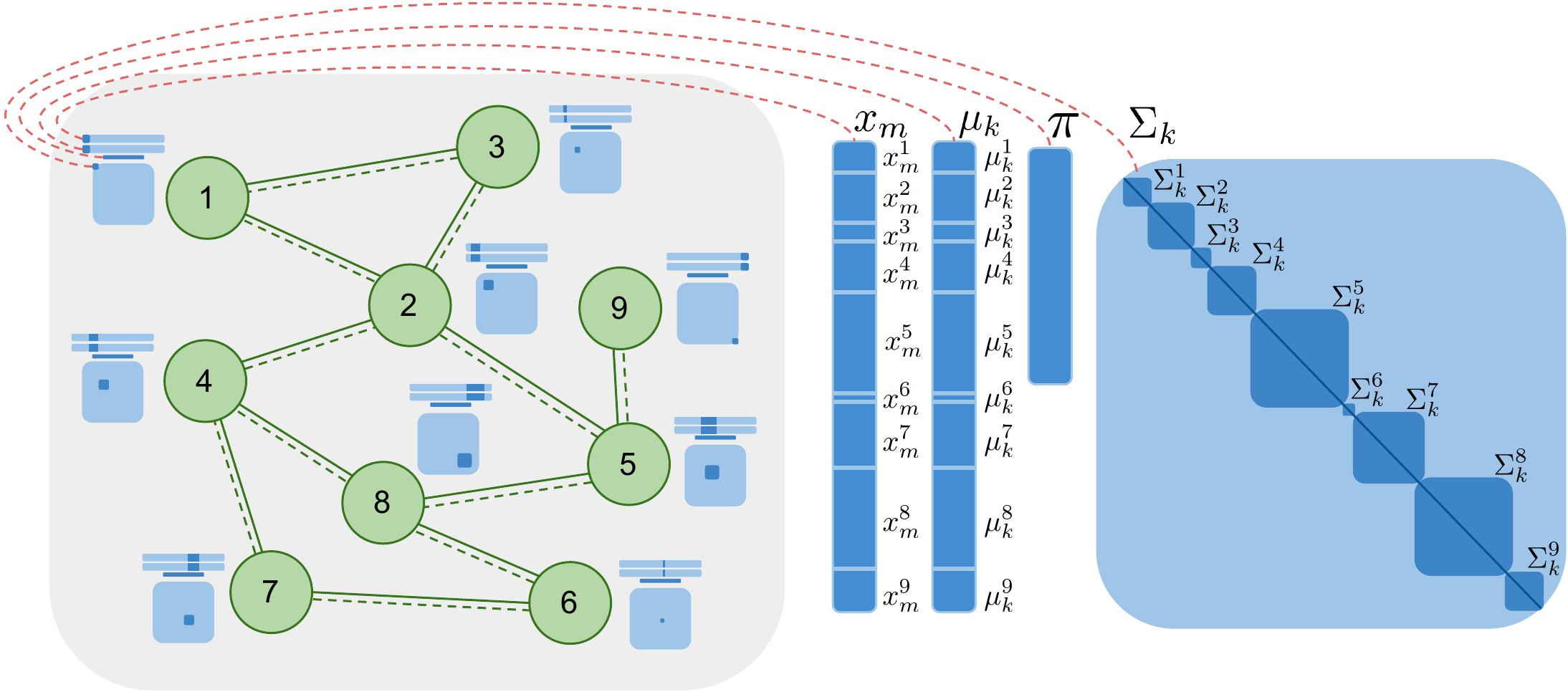}
	\caption{Diagram representing the communication graph of the
		network, as well as the partitioning of the examples and
		parameters in a $0$-hop communication scheme.
	}
	\label{fig:0hop1hopnw}
\end{figure}

In Figure~\ref{fig:0hop1hopnw} we present a network with $N=9$ agents.

\begin{figure}[h]
	\centering
	\includegraphics[width=0.75\columnwidth]{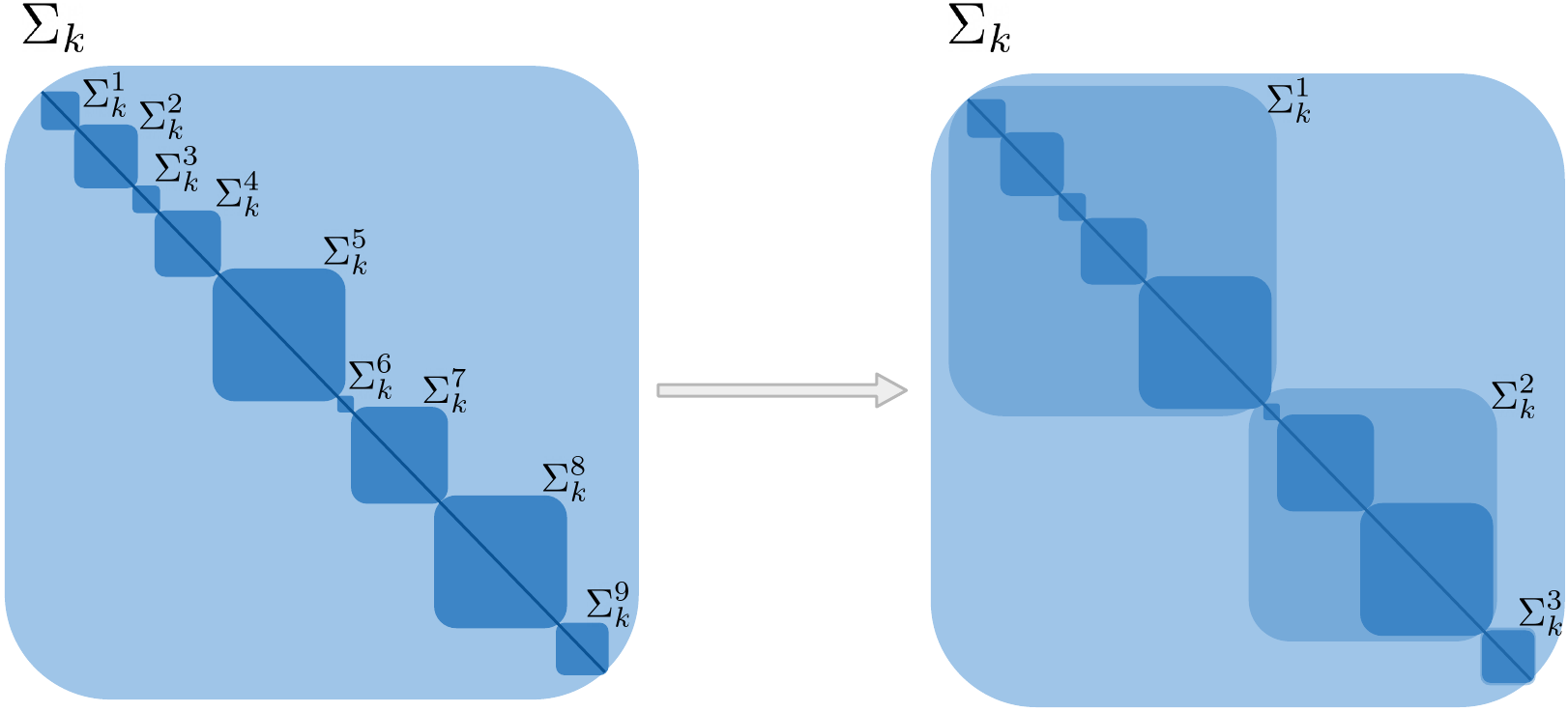}
	\caption{Impact of allowing for 1-hop data sharing:
		going from sparser $\Sigma_k$ matrices estimated in a
		$0$-hop communication scheme to the going less
		sparse $\Sigma_k$ matrices estimated in a
		$1$-hop communication scheme.
		(In the $\Sigma_k$ on the right, the third hub matches the
		partition associated with agent 9 in the $0$-hop scheme.)
	}
	\label{fig:0hop1hop}
\end{figure}

Running the graph clustering algorithm proposed in the paper
for a $1$-hop communication scheme
would lead to hubs $\mathcal{H}_1=\{1,2,3,4,5\}$, with root$(\mathcal{H}_1)=2$; $\mathcal{H}_2=\{6,7,8\}$,
with root$(\mathcal{H}_2)=6$; and
$\mathcal{H}_3=\{9\}$, with root$(\mathcal{H}_3)=9$.
This would, in turn, lead to a different partitioning for
the matrices $\Sigma_k$. A diagram contrasting
the sparsity of $\Sigma_k$ for the $0$-hop and
the $1$-hop schemes is presented in
Figure~\ref{fig:0hop1hop}.

Note that the blocks
$\Sigma_k^1,\dots,\Sigma_k^n,\dots,\Sigma_k^N$ on the left
of Figure~\ref{fig:0hop1hop} ($0$-hop), where every agent
is the root of a hub,
and the blocks $\Sigma_k^1,\dots,\Sigma_k^b,\dots,\Sigma_k^B$
on the right of Figure~\ref{fig:0hop1hop} ($1$-hop),
are the blocks in each root agent $1$-hop.
Therefore, despite having a similar notation,
the blocks $\Sigma_k^1$ on a $0$-hop or a $1$-hop
communication schemes correspond to different blocks.

\bibliography{aistats}